# Retrieval-augmented generation for natural language art provenance searches in the Getty Provenance Index

Mathew Henrickson[1,*], Eric Atwell[1], John Stell[1], Mark Westgarth[2], Dibyadyuti Roy[2], Noorhan Abbas[1]

## Abstract

This study presents a prototype Retrieval Augmented Generation (RAG) framework for art provenance research, focusing on the Getty Provenance Index German Sales dataset. The prototype addresses challenges posed by fragmented and multilingual archival data, as well as the limitations of traditional metadata-based search tools. By enabling flexible, natural language queries in multiple languages, the framework facilitates searches of the Getty Provenance Index without knowledge of specific object metadata. Using a sample of 10,000 records to test the concept and later an extended 100,000 record sample, we explore a RAG prototype that aims to improve both the efficiency and accessibility of provenance searches and find encouraging results for specific and exploratory research scenarios. The framework emphasises transparency, suggesting a scalable and practically oriented approach for historians and cultural heritage professionals working with complex art market archives.

## 1. Introduction

The provenance of an artwork is the 'ownership history of a work of art' [1] and there are several digital resources available for conducting provenance research. One of the leading tools is the Getty Provenance Index (GPI), which is noted as addressing significant challenges in the availability of digital data to researchers [2]. Provenance research is essential for understanding the historical ownership of cultural objects. Active areas of research are those objects affected by Nazi looting [3] [4], and the GPI—German Sales data is a central source for provenance searches linked to this period. Research on German Sales data within the GPI often involves artworks seized during the Nazi period, especially those taken from Jewish families. To confront difficult histories and ties to troubling events in both personal and national stories, it is essential to offer frameworks that are both practical and easy for end users to understand. The problem we looked to address is how the latest search technology can make information retrieval from such resources more efficient and accessible for researchers, while ensuring transparency for the users of the system. The problem was addressed by applying RAG to facilitate natural language access to a corpus of provenance research historic fine art auction catalogues. A similar application has been successful in other domains, for example, in education, to enable students to query textbooks and lecture transcripts [4] and to help lecturers in grading student assessments [5].

### 1.1. The Getty Provenance Index and the role of RAG in modern provenance research

The GPI is a historical art market archive that is widely used by art historians. The GPI in its entirety covers c. 300 years of art market records and text across several different languages. This includes over 12 million art market records, dealer stock books, and archival inventories. This study focuses on the use of the sales catalogue subset of data covering the early-20th century German language art markets. This data collection is used by both art historians and cultural economists researching networks of dealers and art market dynamics [6]. Another significant use of the GPI is in provenance research with a specific focus on artworks looted during the Nazi era [7]. The GPI has become a cornerstone for provenance research, particularly in the context of Nazi-era art sales and restitution efforts. Jacques Schuhmacher (Executive Director of Provenance Research at the Art Institute of Chicago) highlighted the significance of digitised records in provenance research, with particular emphasis on the impact of the GPI—German Sales on provenance searches [8]. These remarks, however, were made within the context of the limitations associated with using online web search portals for locating provenance records. In this study, we build upon these acknowledged advancements by testing a RAG prototype and introducing a novel application of semantic search techniques to further enhance researchers' access to data. Fuhrmeister and Hopp [3] further argued that provenance research must now contend with vast, multilingual, and fragmented data ecosystems, and that scalable, interdisciplinary tools are essential to meet the demands of restitution, transparency, and historical accountability. By combining semantic retrieval with generative summarisation, RAG enables researchers to query large corpora using natural language searches to reveal relevant records and generate explainable summaries grounded in a given context [9]. This is particularly valuable when metadata is incomplete because RAG can infer relevance from embedded semantic cues rather than relying on rigid keyword matching. The application of this technique is designed to enable more flexible and efficient retrieval of provenance information from the GPI. Researchers face the central issue of tracing specific object histories and having to search databases that only handle queries where specific object metadata is often required. A technique that can handle exploratory and more thematic queries could significantly improve this.

[1]School of Computer Science, University of Leeds, Leeds, UK.
[2]School of Fine Art, History of Art and Cultural Studies, University of Leeds, Leeds, UK.
*email: fhmbh@leeds.ac.uk

#### 1.1.1. Objectives and study scope

To summarise, our objectives for the RAG prototype are the following:

- Enable flexible, natural-language queries, such as 'find me paintings records of paintings by [artist name] that contain motifs of family and social activities,' without requiring precise metadata knowledge.
- Support multilingual semantic search, for example, enabling non-German language queries to retrieve relevant content from German-language archives.
- Incorporate semantic-aware retrieval, where the given search terms are automatically expanded to include related or synonymous concepts—e.g., a query for Porträt also retrieves results featuring Mannerbildnis (male portrait) or Bildnis (portrait—synonym).
- Introduce a prototype application of RAG to provenance research to tackle the documented issues of data access and time-consuming archival searches.

### 1.2. Research scope and contribution

This study presents a domain-specialist application and evaluation of a Retrieval-Augmented Generation (RAG) workflow for art-provenance research in the Getty Provenance Index, with outputs designed to support transparent expert verification rather than to produce an automated 'answer'. The work is not framed as proposing a new retrieval architecture or a novel foundation-model technique; instead, it documents how a RAG-style workflow behaves in this specialist setting and reports practical value for provenance research, including the user feedback presented later in this paper. This positioning is aligned with Holocaust-era restitution principles that emphasise transparency and access as practical preconditions for provenance work. The Washington Conference Principles from 1998 [10] call for relevant records and archives to be open and accessible and encourage publicising information to support 'just and fair' solutions, while the Terezín Declaration from 2009 [11] reiterates broad archival accessibility. Considering this important policy backdrop, our study evaluates whether recent advances in language and retrieval technologies can improve the accessibility and usability of large historical art-market sources for provenance search.

#### 1.2.1. Auction catalogue data challenges

Historical auction-catalogue records present domain-specific challenges for information retrieval that differ from conventional 'structured documents'. Entries are typically short and telegraphic, compressing multiple attributes (attribution, medium, dimensions, motifs, and descriptive cues) into catalogue shorthand rather than full sentences. Because there is little redundancy, small variations in wording can significantly affect whether a record is discoverable via search. A further difficulty is the variability of conventions across catalogues. The same attribute may be expressed in multiple notations: dimensions can appear with different separators and ordering (e.g., '44 × 35 cm', '24,5:34,5 cm', 'H 41, B. 32 cm'); media and supports are often abbreviated or inconsistent (e.g., 'Lwd.', 'Leinw.', 'Öl. Leinwand.'); and attributions may be partial or approximate (e.g., 'um 1700', 'Maler des 18. Jahrhunderts'). Catalogues also contain implicit and cross-referential cues (e.g., 'Gegenstück z. Vor.'), which are meaningful to human readers but are difficult to capture through straightforward keyword matching. Finally, auction-catalogue corpora frequently contain multilingual and historically variable entities: artist names, locations, and descriptive terms appear in different languages and variant spellings, and motif descriptions are semantically rich but highly diverse. These characteristics make provenance search less like retrieving a single matching document and more like assembling a reviewable shortlist of plausible candidates from a large pool.

#### 1.2.2. Workflow evaluation

The evaluation is designed to be appropriate for a prototype workflow study in a specialist domain. Following feedback during review, we expanded the query bank from 20 to 100 and evaluated four system configurations (400 total query–configuration runs), added a reranking stage, and increased the indexed dataset from 10k to 100k records. The query set was designed in consultation with domain experts to reflect common provenance search practices and query types. As the evaluation focuses on comparative retrieval-and-shortlisting behaviour across workflow configurations, established IR evaluation work on topic-set reliability informed our decision to use a 100-query set here, consistent with published guidance that 'a good experiment is at least 25 and 50 is better' and that '100 queries is a good target number' for precision-at-k style evaluation [12]. Assessing results also entails careful manual checking of whether returned records are plausible for the researcher's task. Larger-scale benchmarking is therefore a natural extension but is best pursued as follow-on work.

#### 1.2.3. Comparisons with other LLMs

The contribution of this study lies in the domain-specific evaluation of a retrieval-and-shortlisting workflow for provenance search in the GPI, including comparative testing across configurations tailored to typical provenance query types and expert verification. A multi-LLM comparison would change the scope of the work and shift it from a domain-methods and workflow paper into a foundation-model comparison study. In addition, cross-provider comparisons introduce further system-level variables, including prompt sensitivity and provider-specific safeguarding mechanisms, which would require a separate experimental design to control and interpret fairly [13,14]. Cross-provider comparisons also involve additional differences across systems (interfaces and integration choices, context limits, guardrails, and associated retrieval components). For that reason, we treat cross-model benchmarking as a future research direction, while keeping the present contribution focused on documenting and evaluating the workflow under clearly described configurations for this specialist corpus and use case.

#### 1.2.4. Study limitations and future work

We acknowledge a limitation of the present prototype: the embedding and LLM stages rely on closed-source services, although open components are used where appropriate within the workflow. Extending the framework to fully open models and conducting cross-model comparisons is a current priority for further work, and we plan to pursue this through separate evaluation case studies benchmarked against the initial closed-source pipeline prototype.

### 1.3. Related work

Recent studies have increasingly explored the intersection of AI and cultural heritage and highlighted an increasing focus on the integration of such tools into the Humanities. Gîrbacia [15] showed key trends in the application of AI across heritage domains, emphasising the role of semantic technologies. Shinde et al. [16] provided a systematic review of AI in archival science, highlighting parallels with provenance research in data structuring and retrieval. Bushey [17] investigated visual AI in archival contexts, suggesting opportunities for multimodal extensions of text-based systems. Zou and Lin [18] presented case studies on AI in conservation, underscoring the value of interdisciplinary approaches. Together, these studies underline the relevance of integrating AI-based techniques into information retrieval in the Humanities. This study addresses a gap in the current literature, namely, the application of RAG technology to one of the largest provenance databases available to researchers. This study builds on developments made by the Getty Research Institute in providing large-scale art market datasets for provenance research [19], with a particular focus on Nazi-era provenance. In 2011, the Getty, in collaboration with the University of Heidelberg and the Berlin Art Library, digitised over 3200 auction catalogues. Schuhmacher [8] highlights the value of the GPI, describing it as 'short-circuiting searches that could otherwise take years.' Hopp [20] characterises provenance research as a '*Daueraufgabe*' (ongoing task) and calls for greater infrastructure and interdisciplinary collaboration.

Although digitisation has improved general access, the application of RAG techniques, introduced by Lewis et al. [9] in art market and art historical research, is limited. RAG-based techniques are becoming increasingly present in industry and academia for smarter information retrieval [21], but there is a gap in the application in art historical domains. This study builds on these technologies to propose a RAG prototype provenance research model. In the next section, we continue to discuss our aims and study structure to evaluate RAG for provenance research.

Gao et al. [22] outlined the distinct types of RAG techniques currently available. The techniques evaluated in this study are first Naïve Rag, using simple semantic retrieval and LLM summarisation, and in a second evaluation, we incorporated elements of Advanced Rag to evaluate multilingual reranking and metadata filtering configurations. The addition of further features is intentionally limited to favour ease of conceptual understanding by end users. The study, owing to its interdisciplinary application, targets a simpler implementation than some of the latest architectures (see Self-RAG Asai et al., [23], Adaptive RAG Jeong et al., [24]). This is a key consideration because potential future users of the search capability must be able to conceptually understand the pipeline and search mechanics. AI solutions can be powerful but suffer from a 'black box' effect where end users tend to be sceptical of end results of how they work. Explainability, interpretability, and understandability [25] are, therefore, paramount when adapting such technologies for a discipline where trust and reliability play a vital role.

## 2. Materials and methods

The following framework is intended as a complementary tool to the established search portals available to researchers. Its aim is to meet the need and fill the gap in tools where natural language searches help overcome search challenges where specific object metadata is missing. As highlighted by Dr. Sandra van Ginhoven (Head of the Getty Provenance Index, 2025), the application of RAG to the GPI 'enhances the discoverability of provenance data and adds ways to engage with the Getty Provenance Index.' Building on this feedback, we aimed to show the potential of RAG to make searches more flexible and efficient, while broadening the potential research audience by introducing multilingual semantic searches for the first time to the GPI dataset.

For the RAG prototype, we chose models from OpenAI—GPT4o and text-embedding-3-large [26] for text encoding and text summarisation. We combined these models with the FAISS open-source vector index [27] and, in a second stage, introduced an open-source reranker model bge-reranker-v2-m3 [28] to evaluate potential performance gains through reranking. We chose OpenAI text-

embedding-3-large for the text encoding owing to its robust performance in granular semantic retrieval and text embedding benchmarks [29]. FAISS was selected as the vector index owing to its extensive use in academic and commercial retrieval systems [27], its support for high-dimensional and large-scale vector searches, and its flexibility in indexing strategies (e.g., flat, IVF, and HNSW). FAISS enables efficient similarity search across millions of vectors [27], making it well-suited for evaluating retrieval performance in domain-specific corpora, such as historical auction catalogues.

To make the search configuration as flexible as possible, we embedded key metadata into the raw auction text records to enrich the original text. By integrating metadata into the textual content, we added essential information, such as sale date, auction house, or catalogue number, while maintaining a single input record for the model. Our goal in enriching records with embedded metadata was to ensure that more of the object description and sale information is made accessible via the embeddings and semantic search. The result of embedding metadata into the auction record is that the information is made accessible via both semantic search and also metadata filtering if required. The aim is to maximise the chances of relevant information retrieval from either the semantic or metadata-based searches. These combinations were evaluated in phase two of our retrieval evaluation. The text records were then embedded and stored in an FAISS vector index and a separate metadata store for flexibility. The model was then evaluated in two main stages. The first focussed on the initial proof of concept and evaluated the pipeline's recall and final output based on a smaller set of 20 queries and an initial random sample of 10,000 auction records. The first stage focussed on evaluating the recall of the first stage raw retrieval [30], without metadata filtering or reranking, and assessed the quality of the GPT4o output based on this simpler implementation. In a second stage, a more detailed evaluation of context retrieval using a larger set of 100 queries and 100,000 indexed auction entries was conducted, introducing additional metrics and search configurations. The second stage of the retrieval evaluation introduced a reranker model and metadata filtering as possible ways of optimising retrieval relevance and precision. At this stage, we also introduced additional metrics (precision at reranked *k*, and normalised discounted cumulative gain [31]).

In both evaluation phases, we evaluated the queries against a candidate pool of expected records retrieved from our database using SQL queries. The proxy set of records was established by translating semantic searches into SQL searches. An example of this is below.

- Semantic search: '*we are looking for paintings by Picasso, Matisse sold at Fischer in 1939 that featured reference to family, children or women*'.
- Corresponding SQL:

```
SELECT *
FROM GPI_RAG_TABLE
WHERE (artist LIKE '%Picasso%' OR artist LIKE '%Matisse%')
  AND auction_house LIKE '%Fischer%'
  AND sale_date BETWEEN '1939-01-01' AND '1939-12-31'
  AND (
     description LIKE '%Familie%'
     OR description LIKE '%Kind%'
     OR description LIKE '%Frau%'
    )
```

The prototype architecture for both a simple semantic search (excluding reranking) and semantic retrieval (including reranking) are outlined below in **Figure 1**, respectively **Figure 2**.

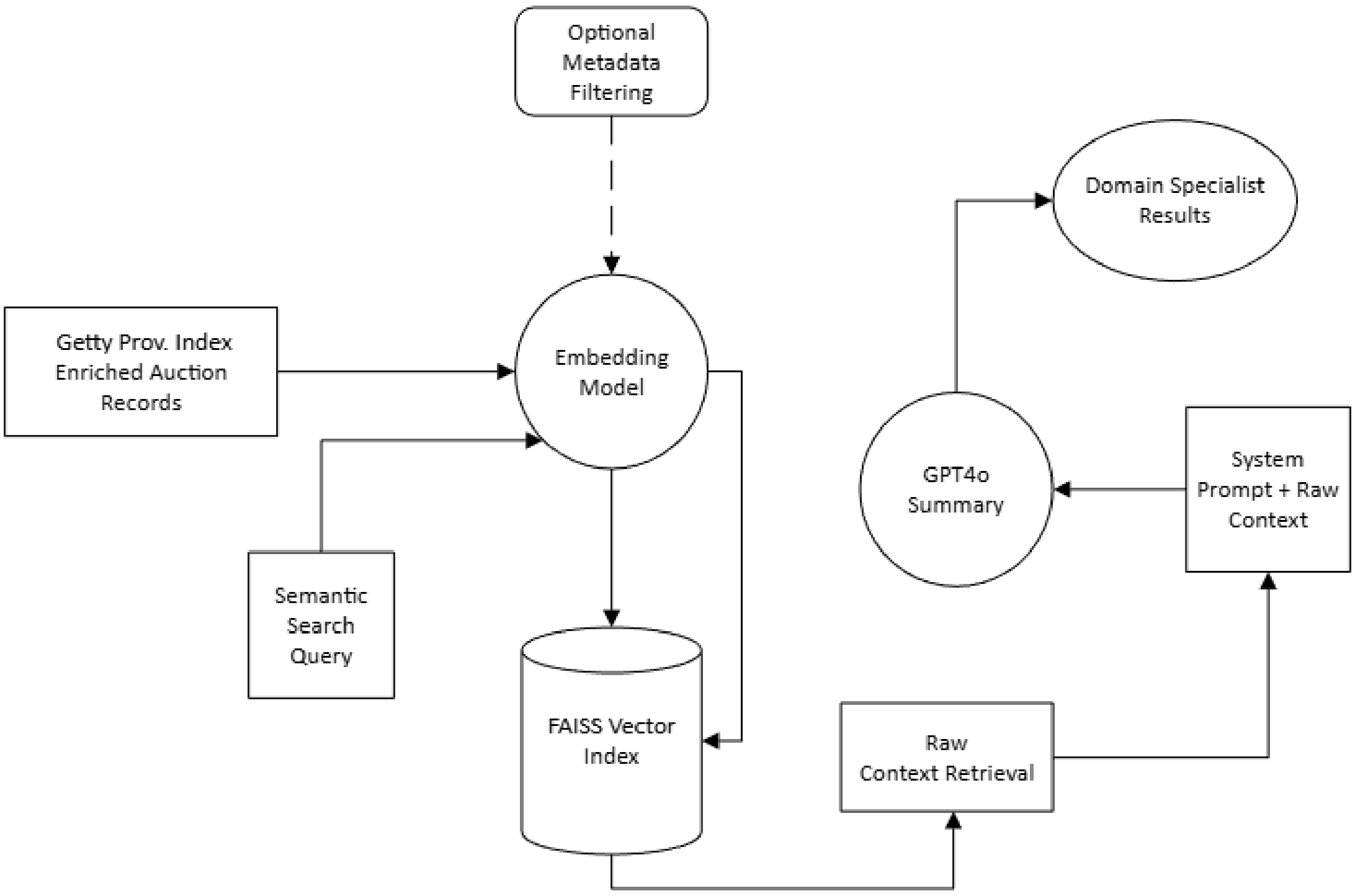

**Figure 1.** An overview of the simple RAG workflow (excluding reranking).

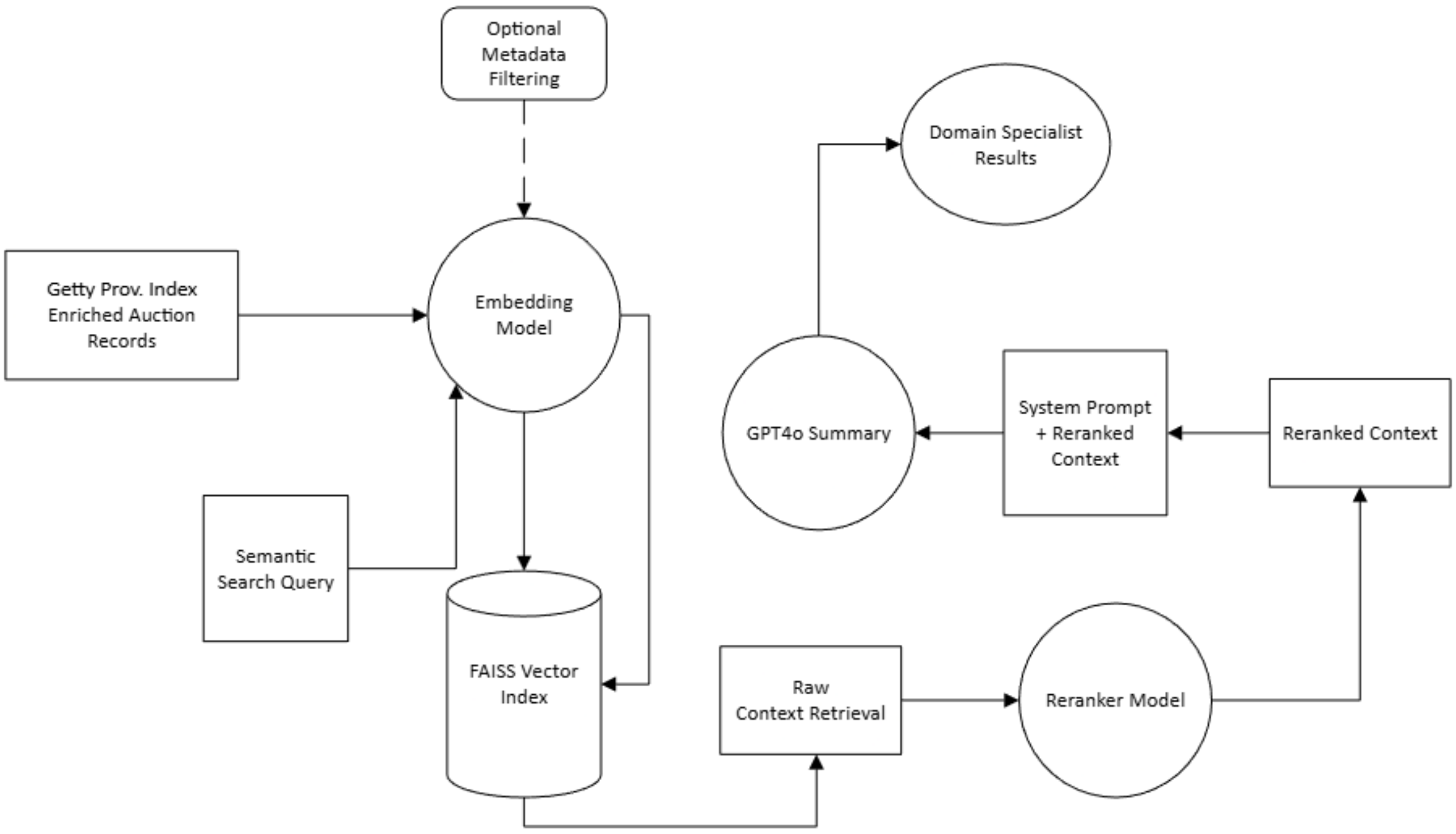

**Figure 2.** An overview of the simple RAG workflow (including reranking).

The stages of our RAG prototype from data preparation and encoding to final retrieval are described in detail below.

- Text Augmentation for Semantic Retrieval. Raw auction catalogue entries were enriched with key metadata fields (artist, object type, auction house, material, dimensions, title or description, auction date, URL link to the original scanned catalogue).

An example of augmented text from the available data is included below:

*Auction House: Fischer Sale Date: 1939-06-30 00:00:00 Artist: Dix, Otto Title: Mutter und Kind. Vor efeuumranktem, dunklem Mauerwerk Kniestück einer frontalsitzenden blonden Frau mit dunkler geöffneter Jacke. Sie hält auf dem Schoss ihren Säugling in zinnoberrotem Strickjäckchen. Rechts oben Ausblick auf blauen Himmel. Signiert rechts unten: O D 1924. Oel auf Leinwand, 76/70 cm. Königsberg/Pr., Städtische Kunstsammlungen. Object Type: Gemälde*

*Metadata: {'source': 'http://digi.ub.uni-heidelberg.de/diglit/fischer1939_06_30', 'sale_date' : '1939-06-30', 'artist': 'Dix, Otto', 'auction_house': 'Fischer' }*

- Text Embedding Generation: The enriched entries were vectorised using OpenAI's text-embedding-3-large model. The model generates 3072-dimensional embeddings that capture nuanced semantic meaning.
- Vector Indexing with FAISS: The generated embeddings were stored in an FAISS index using IndexFlatIP to enable efficient semantic retrieval.
- Query Embedding and Retrieval: For the first evaluation, queries were embedded using the same model and the closest 100 semantic matches retrieved from the index. For the second evaluation phase, the top 100 results were processed with a reranker before being passed to the LLM. In the second evaluation, metadata filtering and reranking were tested.
- Prompt Construction: Retrieved documents were formatted into a structured prompt and given to the LLM for further summarisation. The prompt included a system message that defined the LLM's role and provided clear instructions for summarising the retrieved content. The prompt construction included the context retrieved from the FAISS vector index and the original query for further reference. This was designed to maximise the reliability of the final information retrieved by the user. A URL reference to the original scanned auction catalogues is included in the final response to ensure researchers have a link to the original sales catalogue.

The system prompt (**Figure 3**) and the placeholder from the python code for the context retrieval and initial search query are detailed below.

```
[Prompt Start]
You are a provenance research assistant specializing in historical fine art auction catalogues.
Use **only** the context provided. Do **not** rely on external knowledge or assumptions.
Respond with the rigor and tone of an art historian.
Write all narrative text, explanations, and evaluations **in German**. Maintain the field labels specified below in English
to ensure downstream parsing.
---
### Task:
Classify the research question into one of the following categories:
1. **Object-based** – refers to specific artworks or objects.
2. **General** – examines patterns, trends, actors (e.g., artists, auction houses), or broader provenance contexts.
**Classification Logic**:
- If the question addresses a specific artwork, artist, or small set of objects and structured object data is available,
classify as **object-based**.
- If the question seeks trends, comparisons, or summaries across multiple actors, time periods, or locations, classify as
**general**.
--
### Response Format:
- For **object-based queries**:
- Provide a **complete structured list** of all relevant objects. Use the following fields in exactly this order, keeping the
 English field names:
**Record ID, Artist, Auction House, Material, Dimensions, Description, Location, Provenance Information, Public Source
(URL)**
- **IMPORTANT**: The **Record ID** field is **MANDATORY** for every object. Extract the Record ID from the document header
(e.g., 'Record ID: 582432'). Do **not** invent Record IDs – only use IDs provided in the context.
- Omit other fields if no information is available.
- If only one object is relevant, add a brief explanatory paragraph.
- Clearly mark **Near Matches** if they are semantically similar but not definitively matching.
- If no direct matches are found, list the closest entries and justify this.

- For **general queries**:
- Create a concise summary or bullet points.
- Highlight patterns, actors, or trends.
- Cite at least one URL from the context as evidence.
- List no individual objects unless they support the overall finding.
Add **Provenance Information** when available (e.g., ownership history, collectors, exhibitions, sales). Omit the field if no
 data is available.
Remain factual, avoid speculation, and thoroughly work through the material provided.
--
### Explanation of Inclusion and Exclusion Criteria:
Conclude with a brief paragraph explaining:
- Why each object was included (e.g., direct relevance, semantic proximity, provenance details).
- Why objects were excluded (e.g., lack of relevance, missing key fields, ambiguous assignment).
- How the classification (object-based or general) influenced the selection.
---
### Relevance Assessment:
Evaluate each contextual document (e.g., auction entry, catalog text) regarding its fit to the research question. Use the
following scale:

- **Highly Relevant** – directly answers the question with high factual agreement.
- **Moderately Relevant** – partially suitable; individual metadata missing or semantic connection weaker.
- **Low Relevance** – only minor or tangential connection to the question.
- **Irrelevant** – no discernible connection.
If works by the same artist are excluded, justify this explicitly.
For each document, provide:
- the **Assessment**
- a **brief justification** (1–2 sentences)
Present this assessment as a **numbered list** at the end of the answer and evaluate all up to 10 provided documents.
---### Context:
{formatted_context}

### Research Question:
{question}
---
### Answer (based only on the context above):
[Prompt End]
```

**Figure 3.** LLM system prompt.

In the following section, we outline the evaluation method for assessing the quality of the semantic context retrieval and the final generative LLM output.

# 3. Results

### 3.1. Evaluation methodology

The evaluation process was performed in two stages. The initial stage involved 10,000 records and 20 semantic search queries. The first evaluation focussed on the semantic query search without any reranking or metadata filtering to assess the quality of the semantic retrieval in isolation. The second stage expanded the scope to 100,000 records and 100 queries, incorporating a multilingual reranker model and employing metadata filtering as further search configurations. In the second evaluation, additional metrics were introduced, including precision @ k (10) and normalised discounted cumulative gain [31]. As Yu et al. [32] noted, evaluating RAG pipelines is inherently complex owing to their domain-specific nature. No single standard framework is universally applicable, and this challenge was clear in our study. By dividing the evaluation process into an initial prototype feasibility phase and a subsequent phase involving a larger dataset with the introduction of a reranker model, we were able to gain a more comprehensive understanding of the prototype's potential scalability and performance. In both stages, the LLM-generated output was also manually evaluated for relevance to the initial query on a scale of 1–3 (1—irrelevant; 2—partially relevant; 3—highly relevant).

For both stages of the evaluation, we divided the search queries into four categories. The aim of the different types of queries was to test different query structures to see how the different model configurations performed. The types of queries are defined below. Query categories describe the formulation of the input queries, whereas the object-based vs. general distinction is applied at the generation stage and governs how retrieved records are structured in the output.

- Specific—queries that included clear semantic indicators of object type and artist (i.e., *Were there any paintings by Otto Dix sold at Fischer in 1939?/'Charcoal drawings by Max Liebermann that are signed'*)
- Vague—more general queries detailing what the targeted object(s) may look like or possible object features (i.e., Please retrieve any works that are not paintings and depict motifs Venice and are painted in Gouache)
- Multilingual—queries were assessed in Russian and Mandarin as well as English and German (considered to be the main languages of the GPI) to evaluate the model's multilingual capabilities. The multilingual queries are direct translations of queries already present in the main query bank and share identical search intent; they are evaluated separately solely to isolate and assess the system's cross-lingual retrieval behaviour rather than to introduce a new query type.
- Out of Scope/Irrelevant—control questions that had no link to the dataset to ensure no records were retrieved and to evaluate inaccurate model output.

To establish a benchmark for evaluation, we replicated each query using SQL against our database to generate a set of expected results. For specific queries, it was possible to retrieve complete sets of expected records. However, semantically vague queries could only be approximated using keyword searches. This is an acknowledged limitation of the evaluations, particularly for broader queries where some candidate results may not have been captured in the translated SQL searches.

#### 3.1.1. Proof-of-concept evaluation results

The results of the initial evaluation using semantic-only retrieval of 20 queries on a sample dataset of 10,000 records are detailed below in **Table 1**.

**Table 1**. Proof-of-concept evaluation summary (*n* = 20 queries; 10,000-record sample): recall and manual output ratings by query category

| Query category | Number of queries | Average recall (%) | Manual evaluation |
|---|---|---|---|
| Multilingual * | 2 | 100 | 3 |
| Out-of-scope/irrelevant | 3 | 100 | 2.67 |
| Specific | 8 | 85.2 | 2.88 |
| Vague | 7 | 64.3 | 2.29 |

* Multilingual results correspond to translated variants of existing queries and are shown separately for transparency.

The summary statistics indicated potential for provenance searches but also show limitations for less defined queries. Specific queries returned relevant record retrieval. Notably, the model showed an ability to interpret descriptive and material-based cues, such as identifying terracotta sculptures from indirect references like *'Gebrannter Ton,'* (fired clay) indicating that the model can, in some cases, support more nuanced object-level interrogation. Out-of-scope queries were also handled effectively. For instance, the query '*suspended sharks in tanks exhibited at the Tate*'—relating to a 2021 Damien Hirst Exhibition—was correctly identified as irrelevant, and no records were retrieved, showing reasonable domain boundary control in this restricted test set. Similarly, the query for a sculpture depicting a '*balloon dog by Koons*' was filtered out appropriately, with GPT correctly inferring the artist's name and excluding unrelated results. However, the performance on vague queries was less consistent. The query *'a drawing sold at auction attributed to an Italian artist of the 15th century'* returned a painting instead of a drawing, indicating a failure in media-type filtering. Another query seeking '*sculptures sold by the authorities in Berlin*' only partially matched, suggesting limitations in abstracting institutional references. While some vague queries were handled well—such as the retrieval of '*gouache works depicting motifs of Venice*'—the overall recall and rating for this category were lower, highlighting the need for improved generalisation and semantic abstraction in both retrieval and generation stages. In the next section, we describe the second, more detailed evaluation of the context retrieval testing further search configurations including metadata filtering and reranking. The aim of the second phase of evaluation was to test queries at a larger scale and to gain more insight into poorer-performing queries and how they could be improved.

#### 3.1.2. Detailed context retrieval evaluation

In this section, we describe the more detailed evaluation (**Table 2**) of 100 queries and a dataset of 100K records. The evaluation focuses in more detail on the quality of the information retrieval with different search configurations. These configurations and evaluation metrics [33] for each stage are noted below (**Table 2**).

**Table 2**. Detailed retrieval evaluation setup (100 queries; 100,000 records): four configurations varying metadata filtering and reranking, with corresponding metrics.

| Configuration | Metadata filtering | Reranking | Evaluation metrics |
|---|---|---|---|
| Semantic only (S) | No | No | Recall@100, Precision@100, NDCG@10 |
| Semantic + Reranking (S + R) | No | Yes | Recall@10, Precision@10, NDCG@10 |
| Semantic + Metadata filtering (S + Meta) | Yes | No | Recall@100, Precision@100, NDCG@10 |
| Semantic + Metadata filtering + Reranking (S + Meta + R) | Yes | Yes | Recall@10, Precision@10, NDCG@10 |

The above configurations were selected to evaluate how different search settings perform on the different types of queries and at a larger scale. This choice focuses on the incremental introduction of model complexity to assess the performance of the expanded search configurations. We reported precision, recall, and NDCG because each captures a different aspect of retrieval quality. Recall measures how completely the system retrieves relevant records form the candidate SQL results. Precision reflects the proportion of retrieved items that are relevant and is useful for understanding the reliability of the highest-ranked results, which often receive closer attention during research workflows. NDCG evaluates how well the system orders those results, rewarding relevant items that appear early. These metrics allowed us to assess both the coverage and relevance of the ranked output. An overall summary of the average metrics for each model configuration is shown below. In the following charts, we also report precision for search combinations without reranking. However, in those cases, precision is less meaningful because these configurations can produce small candidate result sets, and we evaluate them using a larger top-k threshold of 100.

The overall results (**Figure 4**) showed that introducing a reranker improved retrieval quality. Metadata filtering also increased precision by narrowing the search space. This is, however, a trade-off as filtering out potential candidate records lowers recall because some relevant auction entries may be excluded when the index is restricted. The reduced recall for the S + R and S + Meta + R configurations is expected, as the reranking stage evaluates only the top 10 semantic candidates rather than the top 100, and metadata filtering further constrains the candidate pool. Although the overall average metrics per search configuration suggest that the introduction of metadata filtering and reranking were beneficial, more analysis was required to gain a deeper understanding of how query types responded to different search configurations.

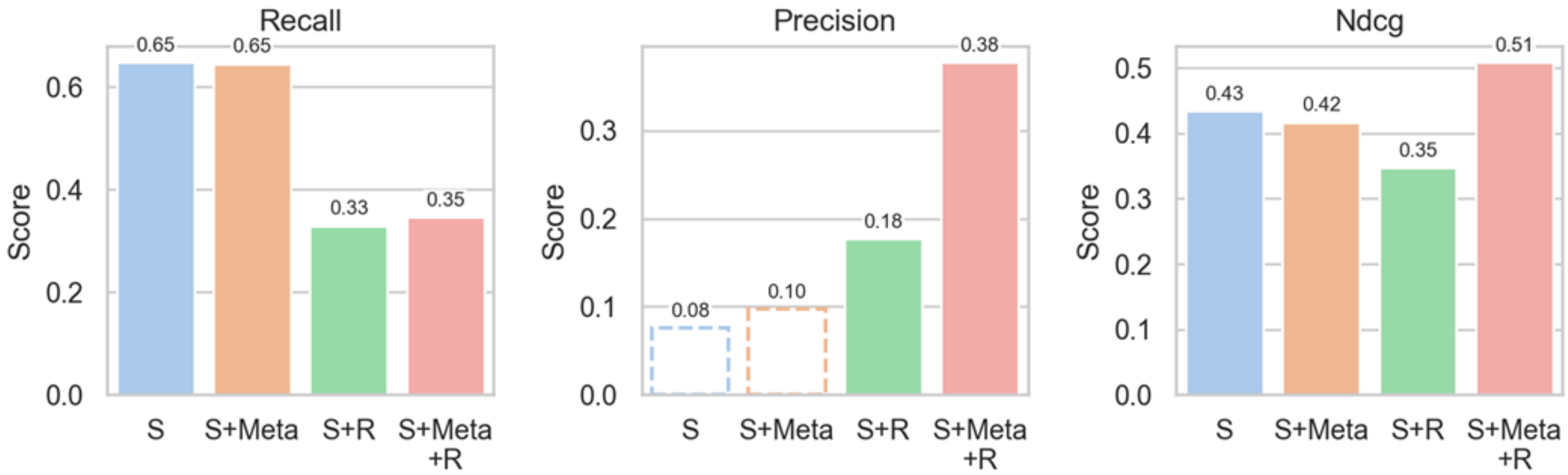

**Figure 4.** Average evaluation metrics per model configuration.

In the following analysis, we present the same metrics but for specific types of queries (**Figure 5**) to highlight how specific queries and vague queries performed using the different search configurations. First, we cover the specific searches.

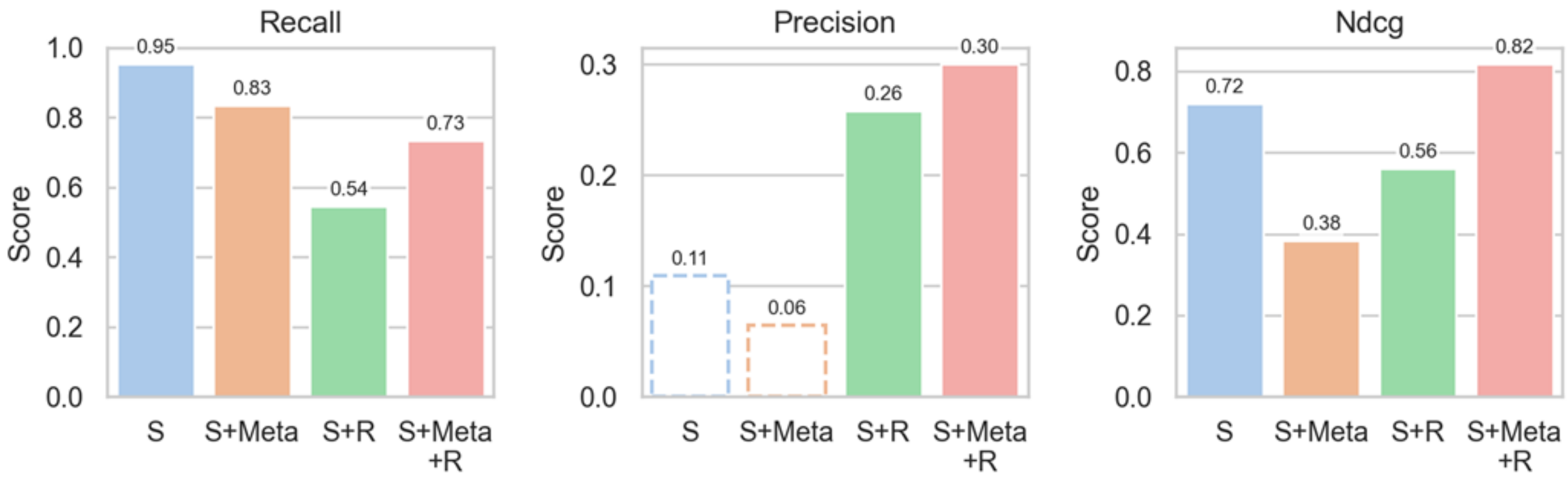

**Figure 5.** Average evaluation metrics—specific queries.

The above chart shows recall, precision, and NDCG for specific queries only. These metrics are consistent with the initial smaller-scale review and confirm that specific queries—those anchored by strong semantic signals such as references to paintings, sketches, studies, drawings, or named artists—perform well across configurations. In this case, there is one optimal setup. The combination of semantic retrieval, metadata filtering, and reranking (S + Meta + R) provides high recall, higher precision, and the highest NDCG, demonstrating that the reranker effectively improves the relevance and ordering of results after filtering the index. It is acknowledged that metadata filtering alone can target some relevant results, but this is never guaranteed, particularly when metadata fields are incomplete or inconsistently populated. Therefore, the combination of semantic retrieval and metadata filtering—followed by reranking—can offer a practical option for researchers. This makes the S + Meta + R configuration especially suitable for specific queries. This evaluation was then repeated for vague queries. The results are detailed in **Figure 6**.

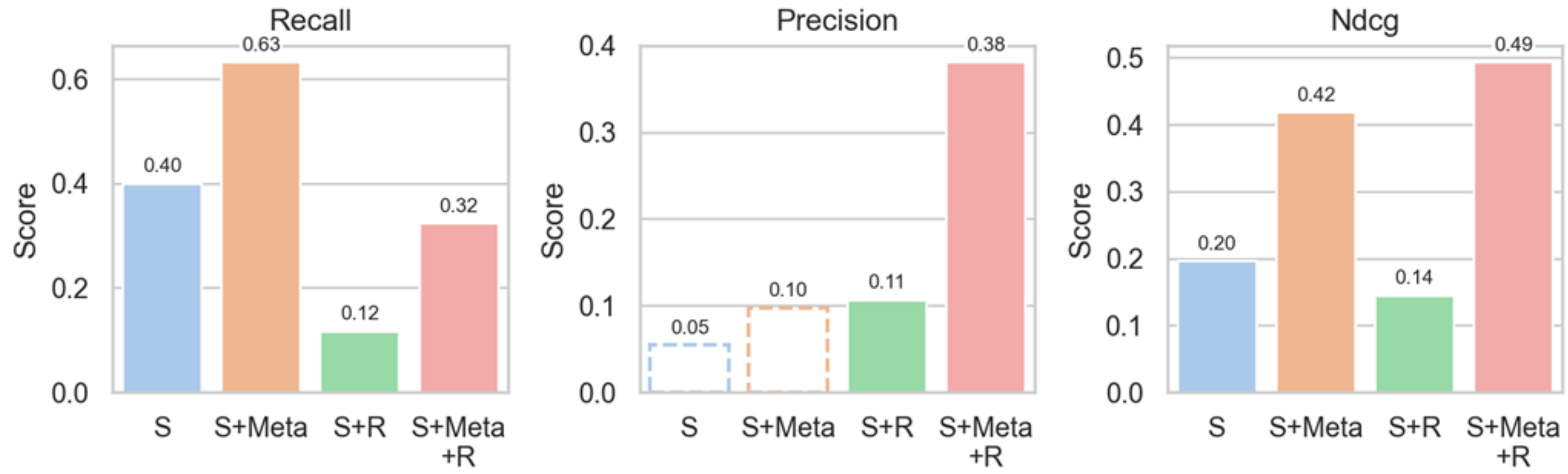

**Figure 6.** Average evaluation metrics—vague queries.

The evaluation of vague queries showed a different pattern from what was initially expected. Because these queries contain weak semantic signals, we anticipated that all configurations would struggle. Semantic-only retrieval (S) does reflect this limitation: recall is lower, and the retrieved items show greater variability, which is consistent with known challenges of embedding-based retrieval when the query does not provide clear discriminative cues. However, when metadata filtering is introduced (S + Meta), the system's behaviour changed noticeably. Even coarse filters—such as restricting the search to a particular auction house or object type—remove a substantial amount of noise from the index. By narrowing the candidate pool, semantic retrieval can operate within a more focussed subset of records, enabling it to pick up motif-level or descriptive overlaps that would otherwise be obscured by the noise in the wider retrieval. This effect was visible in several of our results, where vague queries that lacked explicit semantic anchors nonetheless surfaced entries sharing meaningful descriptive elements. As a result, S + Meta produces higher precision and NDCG than anticipated, and in some cases outperformed configurations without filtering. When reranking is added (S + Meta + R), recall decreases due to the narrower top-k window, but the ordering of results becomes more stable and thematically coherent. In practice, S + Meta provided the broadest coverage for vague queries, while S + Meta + R yielded the most focussed and semantically aligned top-ranked items. Although vague queries inherently cannot achieve full recall—given the lack of clearly defined expected result sets—the combination of metadata filtering and semantic retrieval still enabled the system to identify motif-relevant entries. The examples discussed in the next section demonstrate how the LLM responses highlight semantically related objects retrieved under these broader prompts, despite the queries themselves providing limited structure.

### 3.1.3. Stratified end-to-end evaluation using the best model configuration

A final step in the evaluation involved a second round of manual assessment of the GPT-generated outputs using the best-performing configuration—semantic retrieval combined with metadata filtering and reranking (S + Meta + R). We selected a random sample of 20 queries from the extended query bank and evaluated the resulting LLM outputs using the 1–3 relevance scale. This assessment was carried out with an important limitation in mind: for vague queries, the full set of expected records is often too large to be reproduced in its entirety, and the model was therefore evaluated not on completeness but on the relevance and appropriateness of its final ranked output. Across both specific and vague query types, this final evaluation assessed whether

- The model generates domain-specialist outputs that adhere to the system prompt while remaining faithful to the user's original search intent.
- The model safeguards against irrelevant outputs by restricting results to records that meaningfully align with the query.

The results of the manual ratings are shown in **Table 3**.

**Table 3**. Stratified end-to-end manual evaluation of GPT outputs using the best configuration (S + Meta + R): mean relevance ratings (1–3) by query category ($n$ = 20)

| Query category | Number of queries | Average manual evaluation score (1–3) |
|---|---|---|
| Multilingual | 2 | 3.00 |
| Out-of-scope/irrelevant | 3 | 3.00 |
| Specific | 7 | 3.00 |
| Vague | 7 | 2.43 |

Across the specific queries, performance was consistently strong: the model returned a set of relevant records in every case. A clear illustration of the benefits of combining metadata filtering, semantic search, and reranking comes from the Max Liebermann charcoal drawing queries. Metadata filtering successfully narrowed the pool to Zeichnung (drawings), but the semantic layer identified the correct charcoal works despite highly varied catalogue phrasing. The model consistently recognised medium-specific cues across German formulations such as 'Kohlezeichnung' ('charcoal drawing'), 'Orig.-Kohlezeichnung, mit weiß gehöht' ('original charcoal drawing, heightened with white'), and even longer descriptive lines like 'Aus dem Judenviertel in Amsterdam… sehr interessante flotte Zeichnung' ('From the Jewish quarter in Amsterdam… a very lively sketch'), where the charcoal reference appears mid-sentence. When prompted in English ('charcoal drawings by Max Liebermann'), the system still retrieved and prioritised all these German variants. This shows how semantic matching generalised well to search inputs, while reranking stabilises the final ordering—together producing a reliable, medium-specific result set that metadata alone could not guarantee.

For vague queries, the system's performance was variable; although it struggled with strict precision, it consistently surfaced thematically related records whenever the catalogue text provided even partial motif cues. For example, when asked to 'retrieve works depicting harbour or maritime scenes'—a query without artist, date, or medium constraints—the model returned items described as 'Italienische Meerlandschaft' ('Italian seascape') and 'Seestück' ('sea piece'), correctly identifying the maritime motif despite the underspecified input. Similarly, a broad prompt seeking 'dynamic or action-filled scenes' led the model to surface narrative-rich drawings such as 'Kosack durch einen Fluß reitend' ('Cossack riding through a river') and 'Kosackenvedette am Waldrand' ('Cossack scout at the forest edge'), showing it could infer thematic movement cues from the text. Queries mentioning 'rural or everyday outdoor life' produced outputs like 'Dorflandschaft' ('village landscape') and 'Bäuerin in gebückter Stellung auf dem Felde' ('peasant woman bending in the field'), again demonstrating motif-level semantic alignment. However, the limitations were clear: when asked for 'non-painting works depicting Venetian or lagoon motifs,' the model captured the seascape theme but failed the medium constraint, returning entries such as 'Ölgemälde auf Karton' ('oil painting on cardboard').

In summary, the detailed evaluation illustrated the benefits of including a reranking stage and, where possible, grounding queries with even coarse metadata filters. Although the detailed evaluation showed continuing limitations, we also gained a clearer understanding of some of the issues causing these limitations. This indicates a direction for possible further research and optimisation of the pipeline, detailed in Section 4.1.1. In the next section, we contextualise these results with feedback from the domain of provenance research conducted through user demonstrations and consultations with a German art museum on the wider applications of RAG in museum and art historical archives.

#### 3.1.4. User demonstrations and feedback on RAG for provenance research

To assess whether the RAG pipeline addressed the known challenges of time-consuming searches and inefficient access to relevant information, we presented the initial results to the Getty Provenance Index of the Getty Research Institute (Dr. Sandra van Ginhoven—head of the Getty Provenance Index and her team) and conducted a detailed feedback session with art historian and provenance researcher Dr. Shir Kochavi (Bar-Ilan University). We also discussed the wider implications of RAG systems for museum collections with Jamie Dau and Leslie Zimmermann at the Reiss-Engelhorn-Museum in Mannheim. Feedback from the Getty team was positive, highlighting the potential of RAG to extract information from unstructured catalogue text and to simplify searches for non-experts through natural-language queries. Dr van Ginhoven stated:

> *'The prototype shows incredible potential for provenance data. The ability of users to formulate queries using natural language and of the model to retrieve accurate results that complement those based on existing metadata fields both enhance the discoverability of provenance information and adds ways to engage with Getty Provenance Index data.'*

Dr. Kochavi emphasised the practical benefits of the prototype, explaining that it 'assists in scanning through long lists, highlighting a searchable title or artist,' and stressed how reducing the duration of provenance research can have real impact, as delays can 'postpone potential restitution… causing aggravation to families or Holocaust victims and survivors.' Mr. Zimmermann and Mr. Dau similarly described RAG-based tools as potentially transformative for museums and archives because they lower the technical barriers that often separate experts from the data they curate. This domain expert feedback indicates that, with transparent communication about how RAG works and its limitations, a RAG can meaningfully increase access to provenance data and make search processes more efficient, providing more direct and timely access to relevant candidates within complex archival records.

#### 3.1.5. Workflow comparison vs. the current GPI search portal

The GPI provides an online search portal designed to facilitate the structured exploration of its extensive provenance datasets. Its revamped architecture, grounded in CIDOC CRM Linked Art frameworks, transforms flat-file records into a graph-based network of linked entities, such as artworks, individuals, locations, and events. Linked Art offers a flexible data model that enables consistent, cross-collection discovery by linking cultural heritage records through shared entities and relationships, thereby enhancing usability and interoperability across institutions [34]. The event-focussed model allows researchers to trace meaningful relationships among people, objects, and ideas by modelling events temporally, rather than focussing solely on static object properties [35]. This is useful for tracing intricate relationships between records, actors, and institutions in the dataset. In contrast, the RAG-based prototype introduces a flexible natural language interface that supports exploratory and multilingual queries, semantic abstraction, and contextual summarisation. The main benefit of RAG over the current facility offered by Getty is the flexible and efficient natural language-based search functionality it provides. This removes any technical barriers needed and means that searches can be conducted without precise knowledge of the objects' metadata. In this assessment, we also note that the application of RAG serves a different purpose to the existing tools and is, therefore, to be considered another search technique rather than a competing search technology. Although the RAG solution presents a powerful way of searching provenance datasets, the use of AI models poses some challenges for the discipline relating to transparency traceability. In the next section, we address these concerns for provenance research and the use of AI tools.

# 4. Discussion

### 4.1. AI—RAG and user interpretability

Provenance research encompasses a wide array of historical contexts. Investigations related to German Sales data within the GPI frequently pertain to artworks looted during the Nazi era, particularly those expropriated from Jewish families. To address challenging histories and connections to dark episodes in both personal and national narratives, it is imperative to provide frameworks that are both practical and comprehensible to end users. The framework we have developed emphasises explainable information retrieval (referencing concepts raised by Dwivedi et al.) [36], which is structured to consistently involve a human expert in the process, commonly referred to as a 'human-in-the-loop' workflow. Safdar et al. [37] identified several critical ethical considerations in the use of AI-based applications, notably highlighting the risks associated with 'Blackbox and Bias.' RAG addresses this issue by offering contextually relevant retrieval for an LLM to use in its summaries, thereby reducing the risk of fabricated responses or hallucinations. Additionally, when the context is summarised, we incorporate an evaluation step in the summary that details and justifies the inclusion or exclusion of records from the raw contextual retrieval. Using RAG search will not replace traditional archival search methods and must be presented with adequate disclaimers as to its strengths and limitations. However, what stands out is RAG's ability to offer efficient access to records that would have taken a lot longer to unveil even using existing digital search tools. The following example (**Figure 7**) is an extract from the final user output to show how the records are relatable back to the original online document repositories.

```
[... RAG Output Start]

Input Query: Were there any portraits of men by Rembrandt where the figure was wearing jewellery?

GPT Generated Output:

The research question is classified as object-based because it seeks information about specific artworks by Rembrandt
featuring men wearing jewellery.

Relevant Objects:

Title : Bildnis eines Polen mit hoher goldgeschmückter Samtmütze , unter der das gelockte schwarze Haar
herausquillt . Um den Hals zweireihige Goldkette mit Medaillon .

Artist : Rembrandt

Auction House : Hahn (Heinrich)

Material : Holz

Dimensions : 65 cm x 46 cm

Description : The portrait depicts a man wearing a richly adorned velvet cap and a double gold chain with a medallion around
his neck.

Location : Not specified

Provenance Information : Not provided

Public Source : Hahn 1944 Auction Catalogue

Explanation of Inclusion/Exclusion Criteria:

The item 'Bildnis eines Polen mit hoher goldgeschmückter Samtmütze' was included because it directly matches the query's
criteria: a portrait of a man by Rembrandt where the figure is wearing jewellery (a gold chain with a medallion). Other
items were excluded because of the absence of jewellery in the description or because they were not by Rembrandt. The
classification as object-based influenced the filtering by focusing on specific artworks rather than broader trends or
patterns.

Relevance Evaluation

Highly Relevant – ' Bildnis eines Polen mit hoher goldgeschmückter Samtmütze ‘ by Rembrandt directly answers the
query with a clear description of jewellery .

Irrelevant – 'Männliches Bildnis' by Rembrandt does not mention jewellery .

Irrelevant – 'Brustbild eines Mannes' by Rembrandt does not mention jewellery .

Irrelevant – ' Der Mann mit dem Goldhelm ‘ by Rembrandt does not mention jewellery.

[...RAG Output End]
```

**Figure 7.** Domain specific RAG output.

This format ensures the results are interpretable and traceable to the original context provided by semantic retrieval. This traceability allows end users to evaluate the query results and validate their relevance. The final output was designed to maximise transparency by linking the filtered results to primary historical texts, enabling researchers to quickly reference primary source materials. The framework is specifically designed to address well-documented ethical concerns associated with AI-based tools. Bostrom and Yudkowsky [38] found four key risks in their analysis of domain-specific AI applications: bias and discrimination, lack of transparency, predictability and robustness, and accountability. In response to these risks, we addressed the ethical challenges of bias, transparency, predictability, and accountability through intentional design choices. The structured prompt design and traceable outputs ensure transparency and interpretability, allowing users to understand not only what was retrieved but also why certain records were selected.

In the next section, we outline further research possibilities to explore other model combinations and technical combinations that may result in cheaper and technically better RAG results.

#### 4.1.1. AI—further research

Several alternative strategies exist to improve the model's functionality and potentially improve retrieval accuracy, particularly for more vague queries, where our evaluation identified limitations. Additionally, numerous combinations and RAG implementation options are available for assessment, including retrieval refinement strategies [39]. However, these approaches can become highly technical, necessitating a cautionary note that transparency and traceability for the end user should remain paramount in any model enhancements. Further research could involve fine-tuning smaller LLMs to perform the specific task of provenance search summary rather than relying on larger closed-source models [40, 41]. Moreover, given the limitations in semantic capture from the vague queries, future research could include evaluating the fine tuning of embedding models to the auction text with the aim of improving auction-specific semantic details. The framework could also be expanded to incorporate other art market datasets, such as those hosted by the University of Heidelberg. Integrating digitised data from art market journals of the time could enhance the context provided and allow single searches to retrieve not only relevant auction records from the data but also any references made to relevant artworks in contemporary trade literature.

## 5. Conclusions

This study introduced and evaluated a RAG prototype specifically designed for art provenance research, using Getty Provenance Index (GPI)—German Sales. By enriching auction entry text with metadata and embedding them as extended records, the system enables flexible natural language querying with traceable and explainable outputs. The evaluation results showed robust performance in specific queries, with limitations in vague or abstract searches. These findings underscore the potential of RAG-based systems to support both targeted and exploratory provenance research, while also identifying areas for future refinement in semantic generalisation and media-type filtering. Ethical safeguards are embedded throughout the framework, directly addressing the risks named by Bostrom and Yudkowsky [38] in domain-specific AI applications: bias, transparency, predictability, and accountability. The system design ensures traceable outputs, 'human-in-the-loop' workflows, and contextual grounding in primary archival sources, mitigating the risks of fabricated responses. Rather than replacing existing tools such as the Getty Provenance Index portal, the RAG prototype complements them by offering additional approaches for data and knowledge exploration. It enables researchers to navigate complex historical datasets with greater efficiency while preserving the rigour and contextual sensitivity needed in provenance research. Future iterations may incorporate open-source models and hybrid retrieval strategies; however, the core principles of explainability, transparency, and human oversight must remain central to any such development.

## Acknowledgements

We thank Sandra van Ginhoven (Head, Getty Provenance Index, Getty Research Institute) and Shir Kochavi (art historian and provenance researcher) for domain feedback on the tools and processes developed in this study. The authors acknowledge the use of AI tools as declared in the Materials and methods section. The authors fully support Academia.edu Journals' adherence to COPE guidelines on AI in publication ethics and confirm that this use has been managed responsibly and ethically.

## Data availability statement

The data supporting the findings of this publication can be made available upon request.

## Funding

This research received no external funding.

## Author contributions

Conceptualization, M.H.; methodology, M.H., E.A., and N.A.; software, M.H.; validation, M.H., E.A., and N.A.; formal analysis, M.H., E.A., J.S., and N.A.; investigation, M.H.; M.H.; writing—original draft preparation, M.H.; writing—review and editing, M.H., J.S.,

E.A., D.R., M.W., and N.A.; visualization, M.H.; supervision, E.A., J.S., M.W., D.R., and N.A.; project administration, M.H. All authors have read and agreed to the published version of the manuscript.

## Conflict of interest

The authors declare that they have no competing interests.

## Additional information

## Publisher's note

## Copyright

## References

1. Gerstenblith P. Provenances: real, fake, and questionable. Int J Cult Prop. 2019;26(3):285-304. doi:10.1017/S0940739119000171
2. Sallabedra M. The Getty Research Portal: a decade of progress and developments for the future. Art Libr J. 2024;49(1):2-5. doi:10.1017/alj.2023.27.
3. Fuhrmeister C, Hopp M. Rethinking provenance research. Getty Res J. 2019;(11):213–31. doi: 10.1086/702755
4. Petropoulos J. Art dealer networks in the Third Reich and in the postwar period. J Contemp Hist. 2017;52(3):546-565. doi:10.1177/0022009416637417Alsafari B, Atwell E, Walker A, Callaghan M. Towards effective teaching assistants: From intent-based chatbots to LLM-powered teaching assistants. Nat Lang Process J. 2024;8:100101. doi: 10.1016/j.nlp.2024.100101
5. Abbas N, Atwell E. Cognitive computing with large language models for student assessment feedback. Big Data Cogn Comput. 2025;9:112. doi: 10.3390/bdcc9050112
6. Avery-Quash S, Huemer C, editors. London and the emergence of a European art market, 1780–1820. Los Angeles: Getty Research Institute; 2019. 291p.
7. Gramlich J. Reflections on provenance research: values—politics—art markets. J Art Mark Stud. 2017;1(2). doi:10.23690/jams.v1i2.15
8. Schuhmacher J, De Waal E. Nazi-era provenance of museum collections: a research guide. London: UCL Press in Association with the Victoria and Albert Museum; 2024. 175p.
9. Lewis P, Perez E, Piktus A, Petroni F, Karpukhin V, Goyal N, et al. Retrieval-augmented generation for knowledge-intensive NLP tasks. In: Advances in neural information processing systems. Red Hook (NY): Curran Associates, Inc.; 2020. p. 9459–74 doi:10.5555/3495724.3496517
10. Washington conference principles on nazi-confiscated art. Washington (DC): US Department of State; 1998 [accessed on 14 January 2026]. Available from: https://1997-2001.state.gov/policy_remarks/1998/981203_princ.html
11. Terezín declaration on holocaust era assets and related issues. Prague: Ministry of Foreign Affairs of the Czech Republic; 2009 [accessed on 14 January 2026]. Available from: https://mzv.gov.cz/public/21/14/49/4826349_2940108_Terezin_Declaration_FullText.pdf
12. Buckley C, Voorhees EM. Evaluating evaluation measure stability. Proceedings of the 23rd Annual International ACM SIGIR Conference on Research and Development in Information Retrieval (SIGIR '00); 2000 July 24–28; Athens, Greece. New York: ACM; 2000. p. 33–40. doi: 10.1145/345508.345543
13. Zhuo J, Zhang S, Fang X, Duan H, Lin D, Chen K. ProSA: assessing and understanding the prompt sensitivity of LLMs. In: Findings of the Association for Computational Linguistics: EMNLP 2024. Miami: Association for Computational Linguistics; 2024. p. 1950–76.

14. Dong Y, Mu R, Zhang Y, Sun S, Zhang T, Wu C, et al. Safeguarding large language models: a survey. Artif Intell Rev. 2025;58(12):382. doi: 10.1007/s10462-025-11389-2
15. Gîrbacia F. An analysis of research trends for using artificial intelligence in cultural heritage. Electronics. 2024;13(18):3738. doi: 10.3390/electronics13183738
16. Shinde G, Kirstein T, Ghosh S, Franks P. Tracing the past, predicting the future: a systematic review of AI in archival science. Proc Assoc Inf Sci Technol. 2025;62(1):659–71. doi: 10.1002/pra2.1286
17. Bushey J. Envisioning archival images with artificial intelligence. Archeion. 2024;2024(1):33–54. doi: 10.4467/26581264ARC.24.007.20202
18. Zou D, Lin Z. Research on innovative applications of AI technology in the field of cultural heritage conservation. Acad J Humanit Soc Sci. 2024;7(10) [accessed on 19 November 2025]. Available from: http://francis-press.com/papers/17495#abstract
19. Fredericksen B. The Getty Provenance Index steams ahead. Art Libr J. 1999;24(4):49–51. doi: 10.1017/S0307472200019829
20. Hopp M. Provenienzforschung als disziplin und ihr stellenwert in der wissenschaftslandschaft und universitären Lehre. München: Kunstchronik Monatsschrift für Kunstwissenschaft, Museumswesen und Denkmalpflege; 2021. p. 322–7.
21. Jiang Z, Xu F, Gao L, Sun Z, Liu Q, Dwivedi-Yu J, et al. Active retrieval augmented generation. In: Bouamor H, Pino J, Bali K, editors. Proceedings of the 2023 Conference on Empirical Methods in Natural Language Processing; 6–10 December 2023; Singapore. Singapore: Association for Computational Linguistics; 2023.
22. Gao Y, Xiong Y, Gao X, Jia K, Pan J, Bi Y, et al. Retrieval-augmented generation for large language models: a survey. arXiv. 2024 [accessed on 27 May 2025]. Available from: http://arxiv.org/abs/2312.10997
23. Asai A, Wu Z, Wang Y, Sil A, Hajishirzi H. Self-RAG: Learning to retrieve, generate, and critique through self-reflection. 2023 [accessed on 12 November 2025]. Available from: https://openreview.net/forum?id=hSyW5go0v8
24. Jeong S, Baek J, Cho S, Hwang SJ, Park J. Adaptive-RAG: learning to adapt retrieval-augmented large language models through question complexity. In: Duh K, Gomez H, Bethard S, editors. Proceedings of the 2024 Conference of the North American Chapter of the Association for Computational Linguistics: Human Language Technologies (Volume 1: Long Papers); 16–21 June 2024; Mexico City, Mexico. Stroudsburg (PA): Association for Computational Linguistics; 2024. p. 7036–50 [accessed on 19 November 2025]. Available from: https://aclanthology.org/2024.naacl-long.389/
25. Xu F, Uszkoreit H, Du Y, Fan W, Zhao D, Zhu J. Explainable AI: a brief survey on history, research areas, approaches and challenges. In: Tang J, Kan MY, Zhao D, Li S, Zan H, editors. Natural language processing and Chinese computing. Lecture Notes in Computer Science; Cham: Springer; International Publishing; 2019. vol. 11839, p. 563–74. [accessed on 28 May 2025]. Available from: http://link.springer.com/10.1007/978-3-030-32236-6_51
26. Caspari L, Dastidar KG, Zerhoudi S, Mitrovic J, Granitzer M. Beyond benchmarks: evaluating embedding model similarity for retrieval augmented generation systems. In: CEUR Workshop Proceedings; 2024. Vol 3784. p. 62–70. [accessed on 16 Oct 2025]. Available from: https://ceur-ws.org/Vol-3784/short4.pdf
27. Johnson J, Douze M, Jégou H. Billion-scale similarity search with GPUs. IEEE Trans Big Data. 2017;7(3):535–47. doi: 10.1109/TBDATA.2019.2921572
28. Wang J, Chen K, Chen Z, He P, Zheng W. Winning ClimateCheck: a multi-stage system with BM25, BGE-reranker ensembles, and LLM-based analysis for scientific abstract retrieval. In: Ghosal T, Mayr P, Singh A, Naik A, Rehm G, Freitag D, et al., editors. Proceedings of the Fifth Workshop on Scholarly Document Processing (SDP 2025); 31 July 2025; Vienna, Austria. Stroudsburg (PA): Association for Computational Linguistics; 2025. p. 276–80 [accessed on 3 December 2025]. Available from: https://aclanthology.org/2025.sdp-1.25/
29. Harris N, Butani A, Hashmy S. Enhancing embedding performance through large language model-based text enrichment and rewriting. AAIML. 2024;4(2):2358–68. doi: 10.54364/AAIML.2024.42136
30. Agnihotram G, Sarkar J. Evaluating precision and recall at retrieval time in retrieval-augmented generation (RAG) systems. Am J Comput Sci Technol. 2025;8(4):174–80. doi: 10.11648/j.ajcst.20250804.11
31. Wołk K. Evaluating retrieval-augmented generation variants for clinical decision support: hallucination mitigation and secure on-premises deployment. Electronics. 2025;14(21):4227. doi: 10.3390/electronics14214227
32. Yu H, Gan A, Zhang K, Tong S, Liu Q, Liu Z. Evaluation of retrieval-augmented generation: a survey. In: Zhu W, Xiong H, Cheng X, Cui L, Dou Z, Dong J, et al., editors. Big data. Singapore: Springer Nature; 2025. p. 102–20.
33. Salemi A, Zamani H. Evaluating retrieval quality in retrieval-augmented generation. In: Proceedings of the 47th International ACM SIGIR Conference on Research and Development in Information Retrieval (SIGIR '24). New York (NY): Association for Computing Machinery; 2024. p. 2395-2400. doi:10.1145/3626772.3657957
34. Sanderson R. Implementing linked art in a multi-modal database for cross-collection discovery. Open Libr Humanit. 2024;1(1) [accessed on 20 August 2025]. Available from: https://olh.openlibhums.org/article/id/15407/
35. Bruseker G, Carboni N, Guillem A. Cultural heritage data management: the role of formal ontology and CIDOC CRM. In: Vincent ML, López-Menchero Bendicho VM, Ioannides M, Levy TE, editors. Heritage and archaeology in the digital age. Quantitative Methods in the Humanities and Social Sciences; Cham: Springer International Publishing; 2017. p. 93–131. [accessed on 20 August 2025]. Available from: http://link.springer.com/10.1007/978-3-319-65370-9_6
36. Dwivedi R, Dave D, Naik H, Singhal S, Omer R, Patel P, et al. Explainable AI (XAI): Core ideas, techniques, and solutions. ACM Comput Surv. 2023;55(9):1–33. doi: 10.1145/3561048

37. Safdar NM, Banja JD, Meltzer CC. Ethical considerations in artificial intelligence. Eur J Radiol. 2020;122:108768. doi: 10.1016/j.ejrad.2019.108768
38. Bostrom N, Yudkowsky E. The ethics of artificial intelligence. The Cambridge handbook of artificial intelligence. Cambridge: Cambridge University Press; 2011.
39. Dong J, Fatemi B, Perozzi B, Yang LF, Tsitsulin A. Don't forget to connect! Improving RAG with graph-based reranking. arXiv. 2024 [accessed on 23 September 2025]. Available from: http://arxiv.org/abs/2405.18414
40. Wang L, Yang N, Huang X, Yang L, Majumder R, Wei F. Improving text embeddings with large language models. In: Ku LW, Martins A, Srikumar V, editors. Proceedings of the 62nd Annual Meeting of the Association for Computational Linguistics (Volume 1: Long Papers); 11–16 August 2024; Bangkok, Thailand. Stroudsburg (PA): Association for Computational Linguistics; 2024. p. 11897–916. [accessed on 18 November 2025]. Available from: https://aclanthology.org/2024.acl-long.642/
41. Wang L, Yang N, Huang X, Yang L, Majumder R, Wei F. Multilingual E5 text embeddings: A technical report. arXiv. 2024 [accessed on 27 May 2025]. Available from: http://arxiv.org/abs/2402.05672